\documentclass[10pt,twocolumn,letterpaper]{article}

\usepackage[pagenumbers]{cvpr} %

\usepackage[all]{nowidow}
\usepackage{graphicx}
\usepackage{amsmath}
\usepackage{amssymb}
\usepackage{booktabs}
\usepackage{dsfont}
\usepackage{multirow}
\usepackage{rotating}
\usepackage[inline]{enumitem}
\usepackage[export]{adjustbox}
\usepackage{titling}
\usepackage{bbm}

\renewcommand{\hat}{\widehat}
\renewcommand{\tilde}{\widetilde}

\usepackage[pagebackref,breaklinks,colorlinks]{hyperref}

\usepackage[capitalize]{cleveref}
\crefname{section}{Sec.}{Secs.}
\Crefname{section}{Section}{Sections}
\Crefname{table}{Table}{Tables}
\crefname{table}{Tab.}{Tabs.}

\def\cvprPaperID{*****} %
\def\confName{CVPR}
\def\confYear{2023}

\begin{document}

\title{Watch Your Steps: Local Image and Scene Editing by Text Instructions}

\author{
Ashkan Mirzaei$^\text{1,2}$~~~~~~~~~~~~Tristan Aumentado-Armstrong$^\text{1,2,4}$~~~~~~~~~~~~~Marcus A. Brubaker$^\text{1,3,4}$ \\
Jonathan Kelly$^\text{2}$~~~~~~Alex Levinshtein$^\text{1}$~~~~~~Konstantinos G. Derpanis$^\text{1,3,4}$~~~~~~Igor Gilitschenski$^\text{2,4}$\\
{ $^\text{1}$Samsung AI Centre Toronto~~$^\text{2}$University of Toronto~~$^\text{3}$York University~~$^\text{4}$Vector Institute for AI }\\
}
\date{}

\twocolumn[{%
\renewcommand\twocolumn[1][]{#1}%
\maketitle
\begin{center}
    \centering
    \captionsetup{type=figure}
    \includegraphics[width=0.99\linewidth]{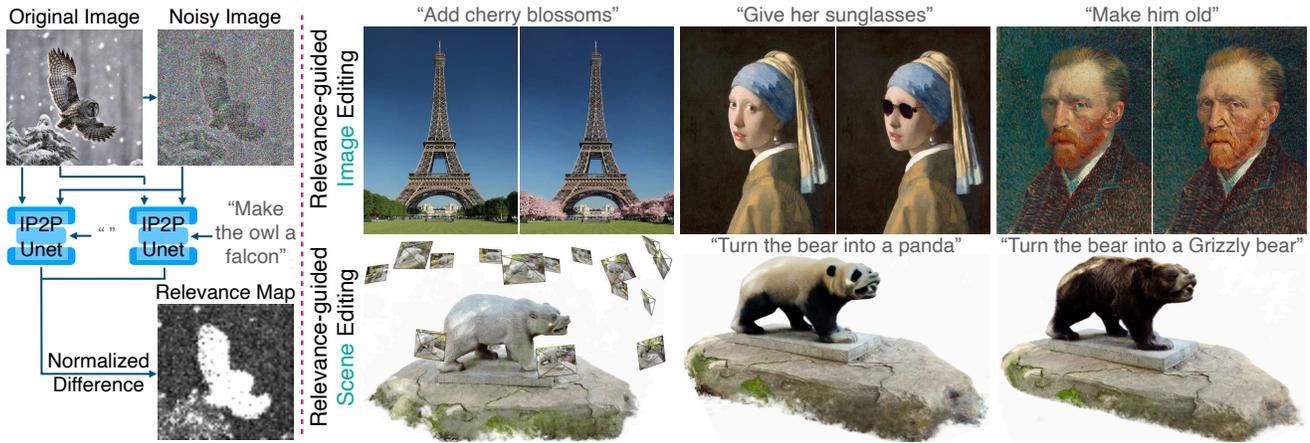}
    \captionof{figure}{Overview of the calculation of the relevance map (left inset), and sample outputs on image (top-right inset) and neural radiance field (bottom-right inset) editing guided by the relevance. Given an image or a Neural Radiance Field (NeRF), our goal is to change the input according to a textual instruction. The relevance map is the disagreement between noise predictions with and without the instruction. For both image and scene editing, we use the relevance map to confine the changes to the most relevant region, according to the edit text. %
    }
    \label{fig:teaser}
\end{center}%
}]

\begin{abstract}
   Denoising diffusion models have enabled high-quality image generation and editing. 
   We present a method to localize the desired edit region implicit in a text instruction. We leverage InstructPix2Pix (IP2P) and identify the discrepancy between IP2P predictions with and without the instruction. This discrepancy is referred to as the relevance map. The relevance map conveys the importance of changing each pixel to achieve the edits, and is used to to guide the modifications. This guidance ensures that the irrelevant pixels remain unchanged. Relevance maps are further used to enhance the quality of text-guided editing of 3D scenes in the form of neural radiance fields. A field is trained on relevance maps of training views, denoted as the relevance field, defining the 3D region within which modifications should be made. We perform iterative updates on the training views guided by rendered relevance maps from the relevance field. Our method achieves state-of-the-art performance on both image and NeRF editing tasks. \href{https://ashmrz.github.io/WatchYourSteps}{Project page}.
\end{abstract}

\section{Introduction}

The crucial role of images in various aspects of modern societies, including social media, marketing, and education, naturally introduces a desire for automated generative approaches for image editing~\cite{couairon2022diffedit,ip2p,meng2022sdedit,hertz2022prompttoprompt,lugmayr2022repaint}. Neural radiance fields~\cite{original.nerf} (NeRFs) are 
increasingly %
accessible~\cite{instant.ngp,mobile.nerf,merf} and popular as an intuitive visualization modality, thus editing NeRFs is also receiving significant attention~\cite{spinnerf,reference.guided.nerf,in2n}. 
The remarkable success of denoising diffusion models~\cite{sohldickstein2015deep,ho2020denoising} in generating high-quality images~\cite{song2020generative,saharia2021image,saharia2022palette,ho2021cascaded,dhariwal2021diffusion} from text~\cite{nichol2022glide,imagen,stable.diffusion,balaji2022eDiff-I} has led to diffusion models being adopted for image editing~\cite{mokady2022nulltext,ramesh2022hierarchical,meng2022sdedit,kawar2023imagic,hertz2022prompttoprompt,Avrahami_2022,couairon2022diffedit,ip2p}. 
Recently, InstructNeRF2NeRF (IN2N)~\cite{in2n} 
demonstrated how to leverage
InstructPix2Pix~\cite{ip2p} (IP2P) for editing NeRFs~\cite{original.nerf}.
We argue that a notable proportion of image and scene editing tasks can be executed by only local modifications. 
Consider the top-right example in Figure~\ref{fig:teaser}, to \textit{``make him old"}. Technically, any output depicting an old man satisfies the instruction. To maintain fidelity to the input and avoid unnecessary variability, it is crucial to confine modifications within local boundaries (within his face in this example)
and naturally prefer parsimonious edits.
By keeping changes within the relevant region, the integrity of the original input is better preserved, while we can ensure that the desired edit is accurately reflected in the output.

Despite the promising results, diffusion-based image editors generally lack a mechanism to automatically localize the edit regions. These methods either ask users for a mask~\cite{lugmayr2022repaint}, rely on the global information kept in a noisy input as a starting point~\cite{meng2022sdedit}, or condition the denoiser on the input~\cite{ip2p}. Nevertheless, all of these methods tend to over-edit~\cite{couairon2022diffedit,ip2p}. Relying on IP2P to iteratively update NeRF's training dataset, IN2N~\cite{in2n} over-edits scenes. 
Recently, DiffEdit~\cite{couairon2022diffedit} proposed using the difference between the noise predictions conditioned on captions to localize image edits; however, it is slow due to \textit{denoising diffusion implicit model}~\cite{ddim} (DDIM) inversion, its quality is inferior to IP2P~\cite{ip2p}, and it requires both input and output captions.

In this paper, we provide an approach to predict the scope implicit in edit instructions to localize image edits. We denote the discrepancy between the noise predictions by IP2P conditioned on the instruction and an empty text as the \textit{relevance map}. Binarizing the relevance map gives the mask of the region that should be edited. We force the denoising process to not change the unmasked pixels by replacing the unmasked region after each denoising iteration with the noisy input~\cite{lugmayr2022repaint}. For NeRF editing by iterative dataset updates~\cite{in2n}, we leverage the power of relevance maps to localize the edits. Note that, across different views, the relevance maps can be slightly inconsistent. To ensure 3D consistency, we train a field on relevance maps from training views, called the \textit{relevance field}. Rendered views from the relevance field are then binarized and used as masks for editing training views. %
Our approach achieves state-of-the-art performance on both image and NeRF editing tasks. 

In summary, (1) we present \textit{relevance maps} to predict the scope of an editing instruction on an image, (2) we use the relevance maps to localize instruction-based image editing, and (3) we lift the maps into 3D by \textit{relevance fields} to leverage the localization in scene editing.

\section{Related work}

\noindent\textbf{Diffusion models for image editing.}
Diffusion models have shown impressive performance in image synthesis~\cite{sohldickstein2015deep,song2020generative,saharia2021image,saharia2022palette,ho2021cascaded,ho2020denoising,dhariwal2021diffusion}. 
Text-to-image diffusion models are able to generate high-quality images based on captions~\cite{nichol2022glide,stable.diffusion,imagen,ramesh2022hierarchical}. Motivated by this success, pre-trained diffusion models have been used to edit images based on text descriptions~\cite{ramesh2022hierarchical,kawar2023imagic,hertz2022prompttoprompt,Avrahami_2022}. SDEdit~\cite{meng2022sdedit} adds noise to input images and denoises them conditioned on a desired description, but lacks a mechanism to keep the details of the original image. DiffEdit~\cite{couairon2022diffedit} uses of the disagreement in predictions of stable diffusion~\cite{stable.diffusion} with input and output captions, but can not handle instructions and fails on more complex captions. 
Recently, IP2P~\cite{ip2p} uses Prompt2Prompt~\cite{hertz2022prompttoprompt} to create a dataset, and trains a denoiser conditioned on edit instructions and the original image. 
IP2P~\cite{ip2p} outperforms the previous methods, but tends to over-edit images.
Simply increasing the image guidance scale or reducing the text guidance scale has adverse effects on the region that actually should be edited. We propose a method to predict the relevance of each latent pixel to the edit, and use it to limit the scope of the edit. 

\begin{figure}[t]
  \centering
   \includegraphics[width=1.0\linewidth]{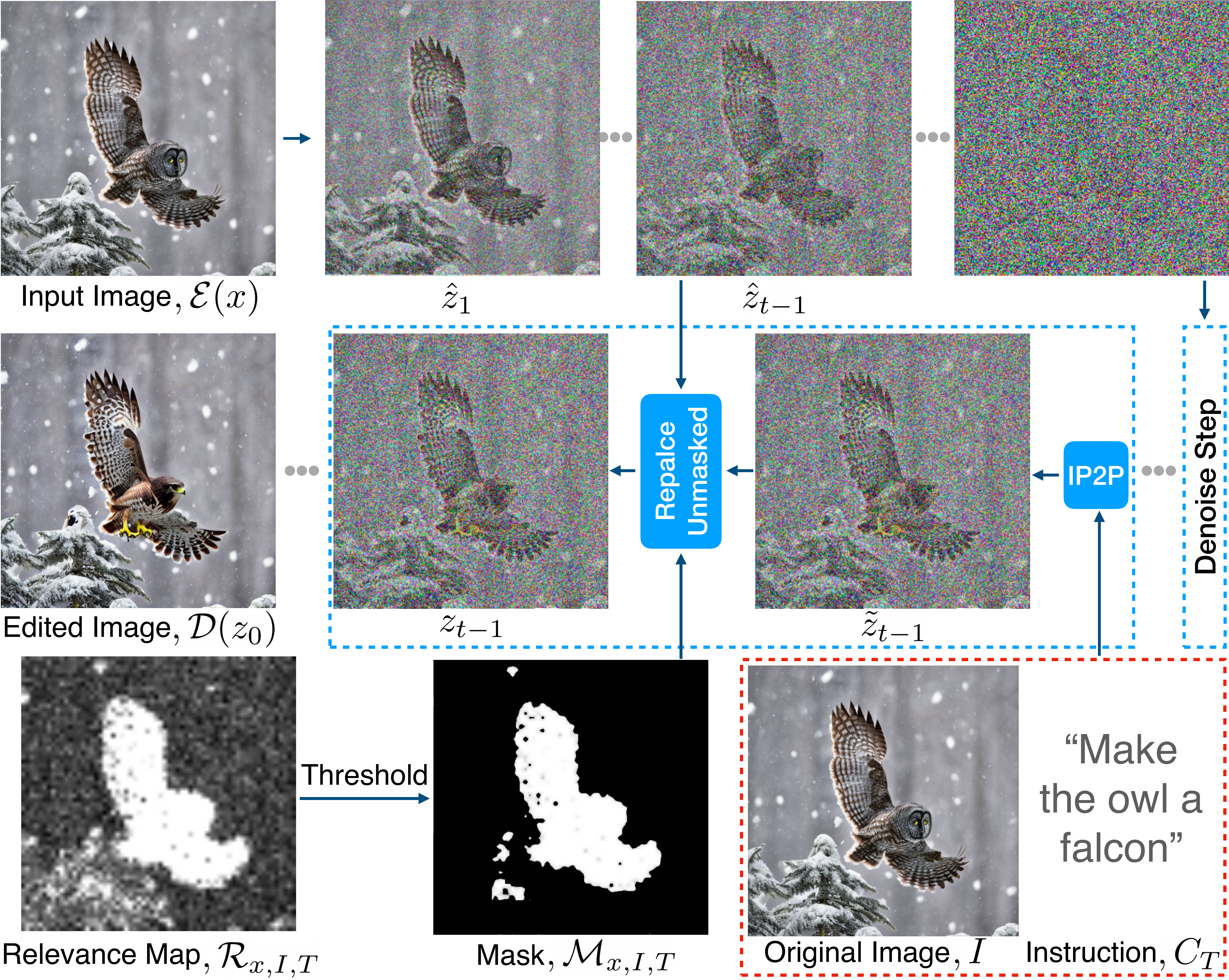}
   \caption{Overview of a denoising step for image editing via relevance-guidance. The relevance map is binarized to get the edit mask. After denoising the output of the last stage with IP2P, the unmasked pixels are swapped with the noisy input to ensure consistency to the input throughout the process. }
   \label{fig:ip2p.local.updates}
\end{figure}

\noindent\textbf{Editing neural fields.}
The advent of NeRFs~\cite{original.nerf} powered by positional encoding~\cite{gehring2017convolutional,vaswani2017attentionisallyouneed,tancik2020fourier} has led to significant popularity of neural rendering models~\cite{advances.in.neural.rendering}. NeRFs are getting faster~\cite{plenoxels,tensorf,instant.ngp,plenoctrees,mobile.nerf,Hedman_2021_ICCV,merf,kurz2022adanerf,barron2023zipnerf, kerbl3Dgaussians}, and less data-intensive~\cite{pixel.nerf,ibrnet,barf,wang2021nerf,poole2022dreamfusion,dreamfields,lin2023magic3d,liu2023zero1to3}, with improved rendering quality~\cite{ds.nerf,mipnerf,mipnerf.360,bacon,refnerf}. The popularity of NeRFs naturally introduces a desire for editing tools. Recent works~\cite{paletteNeRF,laterf,clipnerf,yang2021learning,nerf.editing,liu2021editing,conerf,lazova2023control,song2017semantic,dai2021spsg,jheng2022free,li2022compnvs,mikaeili2023sked} provide NeRF editing approaches, but are typically limited to simple scenes or objects, or perform niche editing tasks. More works~\cite{nerf.in,spinnerf,weder2022removing,reference.guided.nerf} provide 3D inpainting methods to remove unwanted objects from NeRFs. IN2N~\cite{in2n} introduced leveraging IP2P~\cite{ip2p} to edit scenes based on text instructions. Although promising, the outputs lack sharpness due to: i) the spatial ambiguity caused by the IP2P decoder while upsampling latent pixels to patches, and ii) the view-inconsistency due to the 3D-unawareness of IP2P. To alleviate these problems, we leverage our relevance-guided image editor combined with a 3D relevance field to localize the edits in the 3D space and improve consistency.

\section{Background}

\noindent\textbf{Neural radiance fields.} 
NeRFs represent a 3D scene as a neural field, $f_\theta:(x, d) \rightarrow (c, \sigma)$, mapping a 3D coordinate, $x \in \mathbb{R}^3$ and a view direction, $d \in \mathbb{S}^2$, to a colour, $c \in [0, 1]^3$, and a density, $\sigma \in \mathbb{R}^+$. The field parameters, $\theta$, are optimized to fit the field representation to multiview posed image sets. The field is paired with a rendering operator, implemented as the quadrature approximation of the classical volumetric rendering integral~\cite{max2005local}. For a ray, $r$, parametrized as $r = o + td$, where $o$ is the origin and $d$ is the view-direction, rendering begins with sampling $N$ points,  $\{ t_i \}_{i=1}^N$, on the ray between near and far bounds. The rendered colour is then obtained via the volumetric rendering equation, $\hat{C}(r) = \sum_{i=1}^N w_i c_i$, 
where $w_i = T_i(1 - \exp(-\sigma_i \delta_i))$ is the contribution of the $i$-th point, $\delta_i = t_{i + 1} - t_i$ are the adjacent point distances, and $T_i = \exp(-\sum_{j = 1}^{i - 1} \sigma_j \delta_j)$ is the transmittance. 

\noindent\textbf{InstructPix2Pix.}
Given an image, $I$, and a text instruction, $C_T$, describing the edit, IP2P follows the instruction to edit $I$. 
IP2P is trained on a dataset where for each $I$ and $C_T$, a sample edited image, $I_\text{out}$, is given. 
IP2P is based on latent diffusion~\cite{rombach2022highresolution}, where a variational autoencoder~\cite{vae} (VAE) with encoder, $\mathcal{E}$, and decoder, $\mathcal{D}$, is used 
for improved efficiency and quality. 
For training, noise, $\epsilon \sim \mathcal{N}(0, 1)$, is added to $z = \mathcal{E}(I_\text{out})$ to get the noisy latent, $z_t$, where the random timestep, $t \in T$, determines the noise level. 
The denoiser, $\epsilon_\theta$, is initialized with stable diffusion~\cite{rombach2022highresolution} weights, and fine-tuned to minimize the diffusion objective, $\mathbb{E}_{I_\text{out}, I, \epsilon, t} \big[\Vert \epsilon - \epsilon_\theta (z_t, t, I, C_T)) \Vert_2^2\big]$. 
During training, conditions are randomly removed~\cite{liu2023compositional} by setting $I = \emptyset_I$ and $C_T = \emptyset_T$ to further allow unconditional denoising. Thus, the strength of the edit can be controlled by the image guidance scale, $s_I$, and the text guidance scale, $s_T$. The modified score estimate is then obtained as
\begin{align}
    \label{eq:modified.score.estimate}
    \tilde{\epsilon}_\theta (z_t, t, I, &C_T) = \epsilon_\theta(z_t, t, \emptyset_I, \emptyset_T) \notag \\ 
                                               & + s_I \big(\epsilon_\theta(z_t, t, I, \emptyset_T) - \epsilon_\theta(z_t, t, \emptyset_I, \emptyset_T)\big) \notag \\
                                               & + s_T \big(\epsilon_\theta(z_t, t, I, C_T) - \epsilon_\theta(z_t, t, I, \emptyset_T)\big).
\end{align}
After training, the denoiser can be used to either generate edited images from pure noise, or to iteratively denoise a noisy version of an input image to get an output.

\begin{figure}[t]
  \centering
   \includegraphics[width=1.0\linewidth]{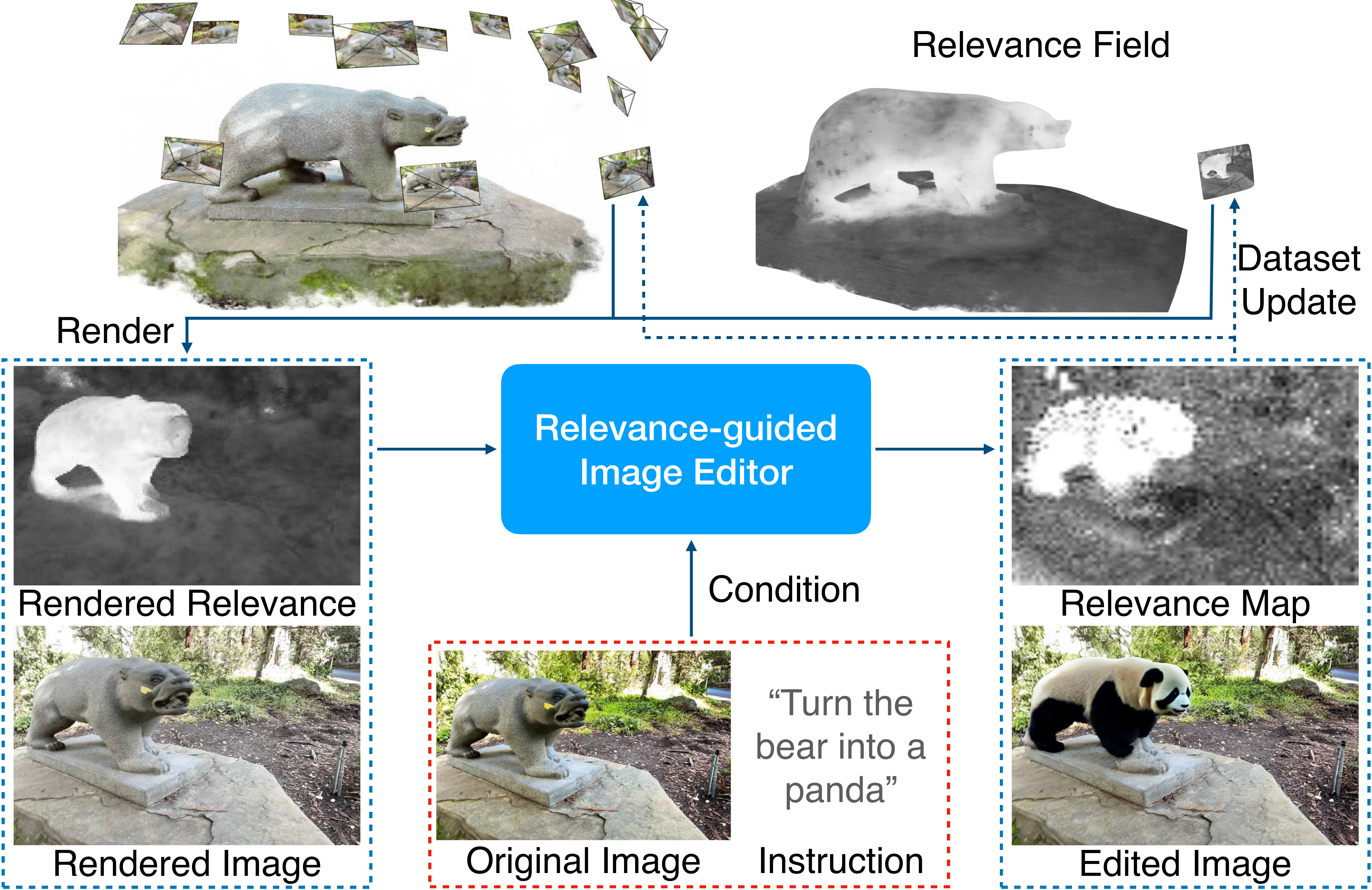}
   \caption{Overview of our relevance-guided NeRF editing method. Iteratively, we take a random view and render it using both the main NeRF and the relevance field. The rendered image is edited guided by the rendered relelvance to only change pixels that are highly relevant to the task. IP2P~\cite{ip2p} is used as the backbone of the editing method, and is always conditioned on the initial captures from the scene. This is to prevent drastic drifts from the original scene in the recurrent synthesis process~\cite{in2n}. The relevance-guided image editing module (\autoref{sec:relevance.guided.image.editing}) returns an edited image and an updated relevance, which are used to update the corresponding training views for the NeRF and the relevance field, respectively. }
   \label{fig:nerf.editing.pipeline}
\end{figure}

\section{Method}

We describe the calculation of the relevance map in~\autoref{sec:relevance.map.calc}. The relevance map is to determine the pixels that should be edited, and is used as a form of mask guidance in~\autoref{sec:relevance.guided.image.editing} for localized image editing. In~\autoref{sec:relevance.field}, we introduce \textit{relevance fields} to allow similar edit localization for 3D scene editing. Implementation details can be found in~\autoref{sec:implementation.details}. 

\subsection{Relevance map calculation}\label{sec:relevance.map.calc}
Given an image, $I$, and an edit instruction, $C_T$, we leverage IP2P~\cite{ip2p} to predict the relevance of each pixel to the edit, i.e., the likelihood that a given pixel needs to be changed, based on the editing task. First, we add noise to the encoded image, $\mathcal{E}(I)$, until a fixed timestep, $t_\text{rel}$, to obtain the noisy latent,
\begin{equation}
    \label{eq:rel.map.add.noise}
    z_{t_\text{rel}} = \sqrt{\alpha_{t_\text{rel}}}\mathcal{E}(I) + \sqrt{1 - \alpha_{t_\text{rel}}}\epsilon,
\end{equation}
where $\epsilon \sim \mathcal{N}(0, 1)$ is a random noise, and $\alpha_t$ is the noise scheduling factor at timestep $t$.
Note that $t_\text{rel}$ is a constant noise level used in our method as a hyperparameter. We then use IP2P's noise prediction Unet, $\epsilon_\theta$, to get two different predictions: i) the predicted noise conditioned on both the image and the text, $\epsilon_{I, T}(z_{t_\text{rel}}) = \epsilon_\theta(z_{t_\text{rel}}, t_\text{rel}, I, C_T)$, and ii) the predicted noise conditioned only on the image and the empty text as the instruction, $\epsilon_{I}(z_{t_\text{rel}}) = \epsilon_\theta(z_{t_\text{rel}}, t_\text{rel}, I, ``")$. 
The difference between $\epsilon_{I, T}(z_{t_\text{rel}})$ and $\epsilon_{I}(z_{t_\text{rel}})$ is that only the former is aware of the text prompt. We use the magnitude of the mismatch between these two values as a measurement of the relevance of each pixel to the edit.
To this end, we first calculate the absolute difference between the two values, which we call the \textit{relevance map},
\begin{equation}
    \label{eq:relevance.map}
    \mathcal{R}_{x, I, T} = \vert \epsilon_{I, T}(z_{t_\text{rel}}) - \epsilon_{I}(z_{t_\text{rel}}) \vert. 
\end{equation}
For robustness, we further clamp the outlier values using interquartile range (IQR) with ratio $1.5$, and normalize the relevance map between 0 and 1. Figure~\ref{fig:teaser} (left inset) illustrates an overview of the calculation of the relevance mask.

\subsection{Relevance-guided image editing} \label{sec:relevance.guided.image.editing}
We propose to use the relevance map to guide the generation of the edited image, and to localize the edited region. For a pixel, a high relevance value means that the pixel is likely to be relevant to the edit, and we allow it to change. In contrast, a low relevance map value signals that the pixel is unlikely to require change.
We apply a mask threshold, $\tau \in [0, 1]$, on the relevance map to get the edit mask, $\mathcal{M}_{x, I, T} = \mathbbm{1}(\mathcal{R}_{x, I, T} \geq \tau)$, enclosing the pixels we allow to be edited. 
To edit an encoded input image, $x$, the encoded image, $\mathcal{E}(x)$, is diffused to a fixed noise level, $t_\text{edit}$, to get the starting noisy latent, $z_{t_\text{edit}}$. The edit noise level, $t_\text{edit}$, determines the strength of the edit; setting it to $0$ results in the input image being unchanged, while setting it to the maximum diffusion timestep starts the generation from pure noise. 
Each denoising stage takes a noisy latent, $z_t$, and denoises it to get $z_{t - 1}$. The denoising step begins with predicting the noise via IP2P to get $\tilde{\epsilon}_t = \tilde{\epsilon}_\theta (z_t, t, I, C_T)$. Using $\tilde{\epsilon}_t$ and the DDIM~\cite{ddim} procedure, the mask-unaware prediction at timestep $t-1$ is
\begin{equation}
    \label{eq:mask.unaware.prediction}
    \tilde{z}_{t - 1} = \sqrt{\alpha_{t - 1}} \Big( \frac{x_t - \sqrt{1-\alpha_t}\tilde{\epsilon}_t}{\sqrt{\alpha_t}} \Big) + \sqrt{1 - \alpha_{t - 1}} \tilde{\epsilon}_t.
\end{equation}
The unedited noisy latent of the input image, $x$, at timestep $t - 1$ would have been $\hat{z}_{t - 1} = \sqrt{\alpha_{t - 1}}\mathcal{E}(x) + \sqrt{1 - \alpha_{t - 1}}\epsilon$.
To obtain $z_{t - 1}$, we combine the mask unaware prediction, $\tilde{z}_{t - 1}$, and the unedited noisy latent, $\hat{z}_{t - 1}$, as
\begin{equation}
    \label{eq:mask.aware.prediction}
    z_{t - 1} = \tilde{z}_{t - 1} \odot \mathcal{M}_{x, I, T} + \hat{z}_{t - 1} \odot (1 - \mathcal{M}_{x, I, T}).
\end{equation}
This way, by replacing the unmasked pixels with the noisy version of the input image, we restrain the generation process from changing any pixel outside of the mask.
After iterative denoising, the edited image, $\mathcal{D}(z_0)$, is obtained. Figure~\ref{fig:ip2p.local.updates} presents an overview of our process.

\subsection{Relevance field for scene editing}\label{sec:relevance.field}
The idea of localizing the edits based on relevance maps can be extended to editing 3D scenes. 
Given a multiview capture, $\{ I_i \}_{i = 1}^n$, of a static scene and the corresponding camera poses, the goal is to edit a NeRF, $f_\theta$, fitted to the scene according to a text prompt, $C_T$. 
Motivated by IN2N~\cite{in2n}, we perform iterative training view updates by replacing one training view, $I_t$, at a time by its edited counterpart according to a text prompt, $C_T$. To ensure the consistency of the localization of edits across different views, we fit a 3D neural field, which we call the \textit{relevance field}, to the relevance maps of all the training views. While editing each of the views, we render the corresponding relevance map from the relevance field to guide the edit.

To implement the relevance field, we extend $f_\theta$ to return a view-independent relevance, $r(x) \in [0, 1]$, for every point, $x$, in the 3D space. Notice that the geometry of the main NeRF and the relevance field is shared, and when fitting the relevance field, we always detach the gradients of the densities to ensure that the potential inconsistencies do not affect the geometry of the main scene. For a ray, $r$, the rendered relevance, $\hat{R}(r)$, can be simply calculated by replacing the point-wise colours with relevance values in the volumetric rendering equation, as $\hat{R}(r) = \sum_{i = 1}^N w_i r_i$.

During the NeRF editing process, every $n_\text{edit}$ iterations, we randomly sample a training view $I_i$. The first time we sample $I_i$, the relevance map, $\mathcal{R}_{I_i, I, T}$, is calculated and added to the training data of the relevance field. From the same view, the image, $\hat{I}_i$, and the relevance map, $\hat{R}_i$, are rendered using $f_\theta$. The relevance-guided image editor from~\autoref{sec:relevance.guided.image.editing} is used to locally edit $\hat{I}_i$, conditioned on the original captured image, $I_i$, and the text condition, $C_T$. To this end, the encoded rendered image, $\mathcal{E}(\hat{I}_i)$, is diffused until a random timestep, $t_\text{edit} \in T$, to obtain $z_{t_\text{edit}}$. The noisy latent, $z_{t_\text{edit}}$, is iteratively denoised conditioned on the original unedited view, $I_i$, and the text prompt, $C_T$, guided by the rendered relevance map, $\hat{R}_i$, to obtain the edited training view, $\tilde{I}_i = \mathcal{D}(z_0)$. 
Since the several-fold upsampling induced by the decoder could lead to inconsistencies in the unedited region,
we replace the unedited RGB pixels in $\tilde{I}_i$ with their counterparts from $I_i$ using a relevance mask rendered in the original image resolution. After editing, $\tilde{I}_i$ replaces the corresponding training view to supervise the main NeRF (the color field).

\subsection{Implementation details}\label{sec:implementation.details} 
In all of our experiments, we set $t_\text{rel} = 0.8$, i.e., we apply $80\%$ of the noise to predict the relevance map.
For IP2P, we used the model available on HuggingFace, based on the diffusers package. For NeRF editing, we used the nerfacto model from NeRFStudio~\cite{nerfstudio}. During the iterative dataset updates, we performed edits with noise levels (timesteps) randomly sampled from $[ 0.02, 0.98 ]$. We update a single training view every $n_\text{edit}=10$ iterations. Each image is updated using $20$ denoising steps for NeRF editing, and $100$ denoising steps for image editing. For the relevance field implementation, we borrowed the same hyperparameters as the nerfacto field~\cite{nerfstudio}; however, we never used the densities from this field, and only used the geometry from the main radiance field. The threshold, $\tau$ is set between $[0.4, 0.6]$ in all the experiments, unless stated otherwise. Each NeRF is first trained for $30{,}000$ iterations on the original scene, and then edited for $3{,}000$ or $4{,}000$ iterations depending on the number of training views.

\section{Experiments}

\noindent\textbf{Datasets.}
For image editing, we follow IP2P~\cite{ip2p} and use their dataset of diverse images and editing instructions. The test set we used consists of $5{,}000$ images, paired with instructions, and input and output captions. NeRF editing evaluation is done using scenes from IN2N~\cite{in2n} and LLFF~\cite{llff}. We use 14 different NeRF editing tasks (i.e., text instructions) for the quantitative experiments. For each, a scene is edited using an instruction, and evaluated against a desired output caption. IN2N and LLFF provide multiview captures of forward-facing and $360^\circ$ static scenes. Colmap~\cite{schoenberger2016sfm,schoenberger2016mvs} is used to recover camera parameters.

\noindent\textbf{Metrics.}
Following IP2P~\cite{ip2p}, for image editing, we use \textit{CLIP image similarity}~\cite{clip.paper} and \textit{CLIP text-image direction similarity}~\cite{gal2021stylegannada}. The former is the similarity between the CLIP embeddings of the edited and the original images. The latter measures the agreement between the change of the images (in CLIP space) and the change of the text captions. 
For scene editing, we use CLIP text-image similarity (Txt-Img Sim.) which is the cosine similarity of the CLIP embeddings of output views and the output caption. In addition, CLIP frame similarity (Frame Sim.) measures the cosine similarity of the consecutive frames in CLIP space. CLIP edit similarity (Edit Sim.) measures the directional agreement between the changes applied to neighbouring frames in CLIP space, as described in IN2N~\cite{in2n}. CLIP image similarity (Image Sim.) and edit PSNR measure the consistency of the edited views and the input views, in the CLIP and RGB spaces, respectively. Finally, we use NIQE~\cite{niqe}, a no-reference image quality metric, to evaluate the quality and sharpness of the outputs.

\noindent\textbf{Image editing baselines.}
For image editing, we compare against state-of-the-art methods, including DiffEdit~\cite{couairon2022diffedit}, SDEdit~\cite{meng2022sdedit}, and IP2P~\cite{ip2p}. Notice that DiffEdit expects input and output captions, and is evaluated with its desired inputs instead of the edit instruction. SDEdit expects the output caption; we evaluate it with the output caption as \textit{SDEdit (out caption)} and with the edit instruction as \textit{SDEdit (instruction)}, separately. 

\noindent\textbf{NeRF editing baselines.}
We quantitatively evaluate against IN2N~\cite{in2n} and per-frame IP2P, which independently edits rendered views of the input NeRF via IP2P~\cite{ip2p}. We further compare our model against NeRF-Art~\cite{nerf.art}, which uses CLIP similarity of the scene and a caption to edit scenes. Additional baselines include IN2N~\cite{in2n} with stable diffusion~\cite{stable.diffusion} (SD) rather than IP2P, and the Score Distillation Sampling~\cite{poole2022dreamfusion} (SDS) loss with IP2P. 

\subsection{Results}

\begin{figure}[t]
  \centering
   \includegraphics[width=1.0\linewidth]{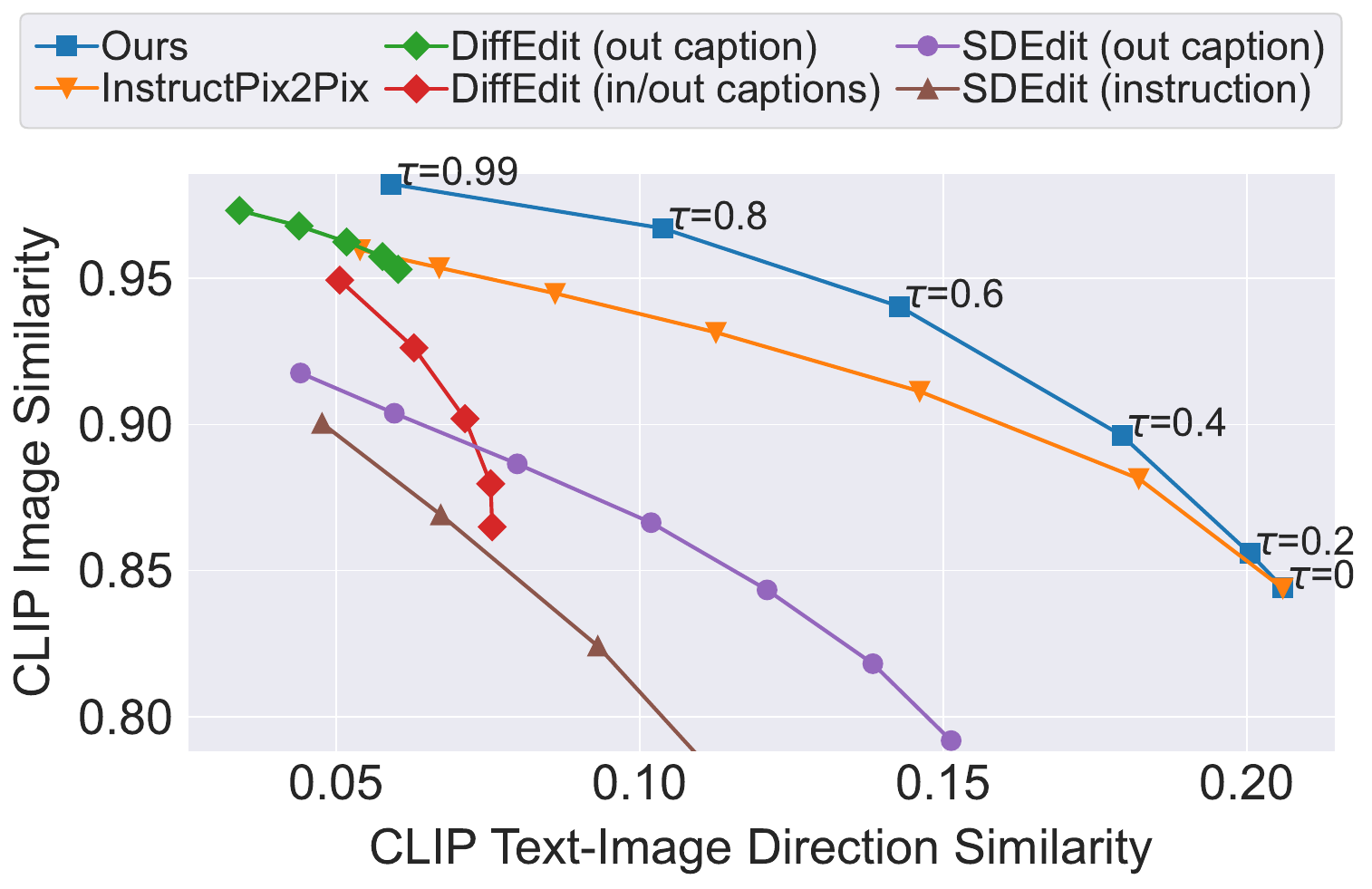}
   \caption{
   Quantitative image editing evaluation. Our model achieves better text-image direction similarity (x-axis), while maintaining higher fidelity to the input
   (y-axis). The text-guidance is set to $7.5$ for every method. We pick SDEdit's strength from $[0.1, 0.9]$ and Diffedit's encoding-ratio from $[0.5, 0.9]$. For IP2P, $S_I$ is changed between $[1, 2.2]$. For our method, $s_I$ is set to $1$.  
   }
   \label{fig:2d.comparison.quantitative}
\end{figure}

\begin{figure*}[t]
  \centering
   \includegraphics[width=1.0\linewidth]{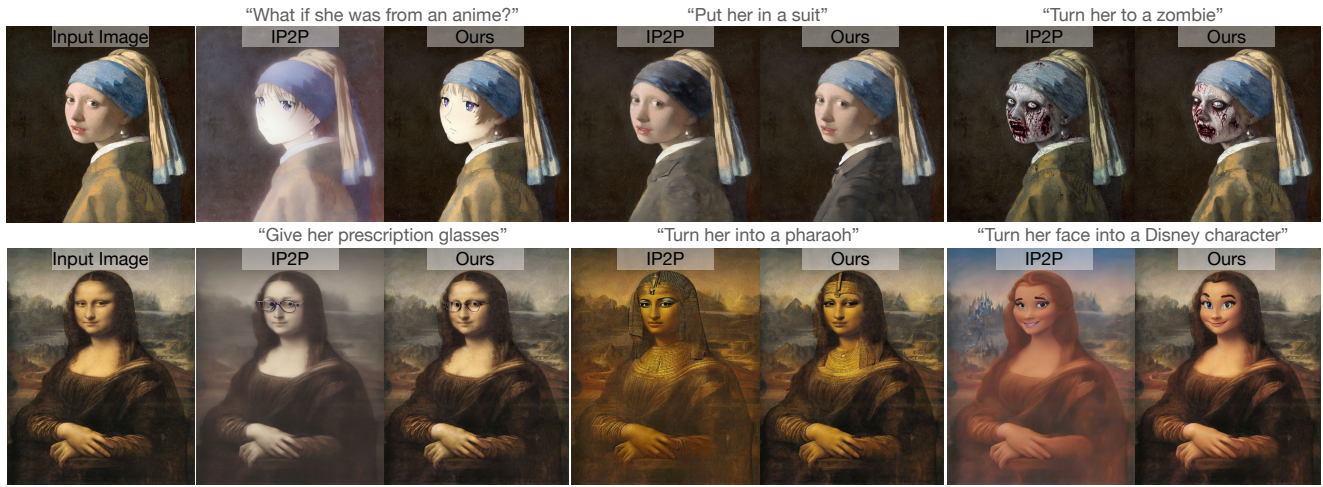}
   \caption{Our image editing results compared to IP2P. For both models, $s_T$ and $s_I$ are set to $7.5$ and $1$, respectively. IP2P fails to isolate the specified region, and over-edits the input. Our model explicitly predicts the scope of the edit, and limits the edit inside a specific region.  }
   \label{fig:more.2d.comparison.ip2p}
\end{figure*}

\begin{figure*}[t]
  \centering
   \includegraphics[width=1.0\linewidth]{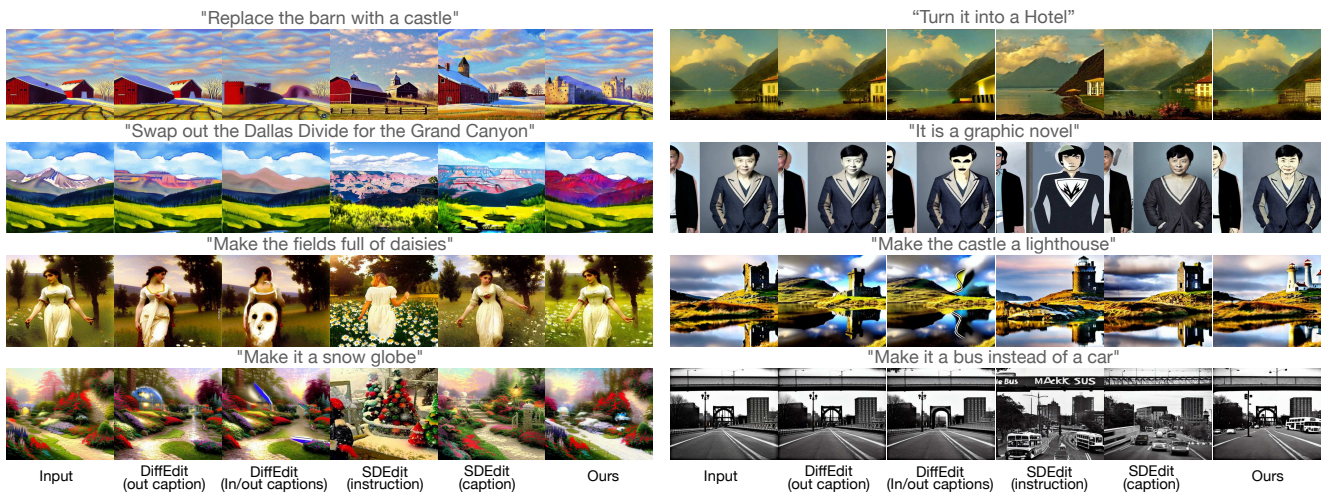}
   \caption{Comparison of our image editing method against DiffEdit~\cite{couairon2022diffedit} and SDEdit~\cite{meng2022sdedit}. DiffEdit requires the captions of both the input and output, but still fails to perform the edit as the captions in IP2P~\cite{ip2p} dataset are relatively complex. SDEdit~\cite{meng2022sdedit} performs better when it is given the output caption. Our model follows the instructions more closely, while maintaining coherence with the input.  }
   \label{fig:multiple.baselines.2d.comparison}
\end{figure*}

\begin{figure*}[t]
  \centering
   \includegraphics[width=1.0\linewidth]{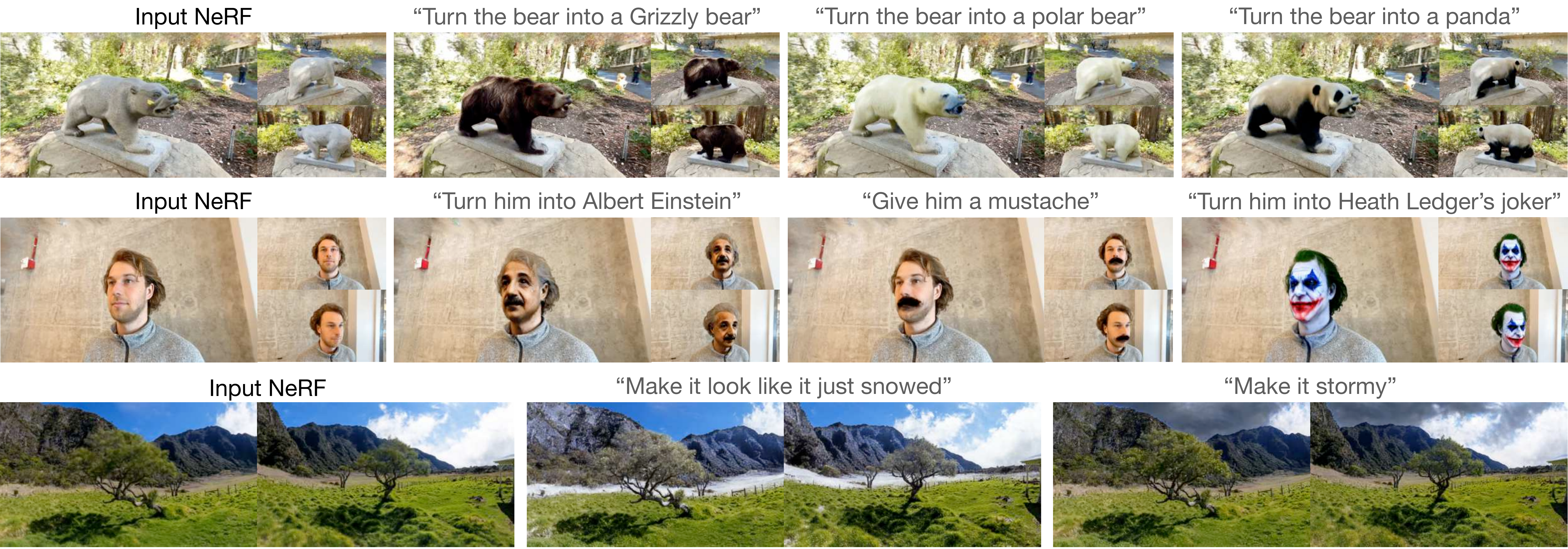}
   \caption{Qualitative outputs of our NeRF editing method. For each of the 3 scenes and each text instruction, we provide multiview renderings of the edited NeRF to show view consistency. Our method follows the text, yet keeps the regions less relevant to the task intact. }
   \label{fig:3d.qualitative.results}
\end{figure*}

\noindent\textbf{Image editing.}
We provide quantitative evaluation of our relevance-guided image editing method in Figure~\ref{fig:2d.comparison.quantitative}, based on the IP2P~\cite{ip2p} dataset. The figure shows the result of each model based on two competing metrics; similarity to the input image (y-axis), and the agreement with the edit (x-axis). Compared to the baselines, our model achieves higher image consistency with similar directional similarities. Additionally, notice that increasing the mask threshold, $\tau$, increases the image similarity as a smaller region of the images is being edited. However, overly increasing $\tau$ has a detrimental effect on successfully achieving the edit. 

DiffEdit~\cite{couairon2022diffedit} requires access to both input and output captions. Even with this information, since the captions in IP2P dataset are relatively complex rather than simple class names or high-level descriptions, DiffEdit fails to  perform appropriate edits. In particular, when DiffEdit is only given the output caption and an empty text as the input caption, i.e., \textit{DiffEdit (out caption)}, it never achieves higher text-image similarities, and the inputs remain relatively unchanged. This is due in part to the failure of DiffEdit in predicting the right masks; thus, the generative process is unable to apply proper changes to achieve successful edits. 
For SDEdit~\cite{meng2022sdedit,stable.diffusion}, the fidelity of the outputs to the input images drop significantly as the strength of the edit is increased. This drop is due to the lack of an explicit mechanism to ensure consistency. In contrast to our model, SDEdit relies on the information kept in the noisy latent; however, in later diffusion stages, the noisy latent retains global information about the input, but lacks local details.

In many editing scenarios, the task can be fulfilled by only changing a local region of the image. Our method outperforms all the baselines by localizing the edits, 
and is able to produce on-par text-image similarities to IP2P, while keeping the outputs more consistent with the inputs.  We provide qualitative comparisons in Figures~\ref{fig:more.2d.comparison.ip2p} and \ref{fig:multiple.baselines.2d.comparison}. Our outputs closely match the inputs; edits are only applied where necessary. See supplemental for more examples.

\noindent\textbf{NeRF editing.}
We evaluate our method against the baselines on scenes from LLFF~\cite{llff} and IN2N~\cite{in2n}. Table~\ref{tab:3d.editing.results} contains quantitative results based on 14 scene editing tasks. Both our method and IN2N perform on-par with per-frame IP2P in terms of CLIP~\cite{clip.paper} text-image similarity, meaning the edited scenes successfully match the output captions. CLIP frame similarity and CLIP edit similarity show that IN2N and our method produce view-consistent results, whereas IP2P independently edits rendered views and is unable to maintain consistency. CLIP image similarity and Edit PSNR compare the cosine similarity and PSNR between each rendered view from the edited NeRF and from the input NeRF. They show that our method keeps the edited scene more consistent with the input scene. Finally, in terms of NIQE~\cite{niqe}, a no-reference image quality metric, our method outperforms IN2N by producing sharper and higher quality results. Since per-frame IP2P's outputs are direct returns of a diffusion model rather than a NeRF, they lack typical NeRF artifacts and thus NIQE is higher for IP2P.

\begin{figure}[t]
  \centering
   \includegraphics[width=1.0\linewidth]{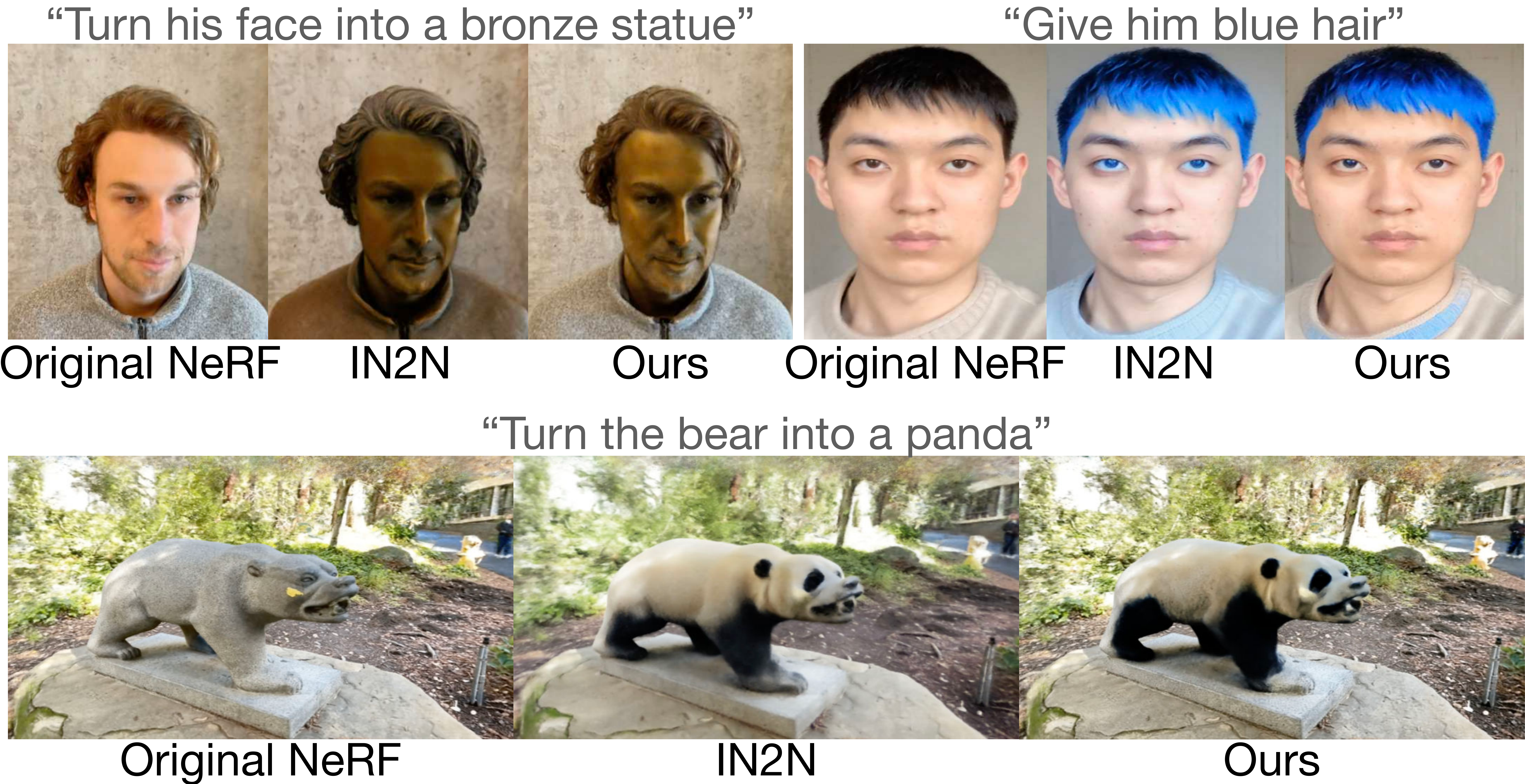}
   \caption{Comparison of our scene editing results against IN2N~\cite{in2n}. 
   The relevance field enables us to localize the edit to the most significant regions.
   Editing a smaller region reduces the decoder spatial ambiguity problem on unedited pixels. Moreover, it improves the view consistency in the edited region as editing a small part is more likely to produce consistent results across the views. Finally, having a small NeRF reconstruction loss outside of the mask improves the convergence in the masked region.  }
   \label{fig:in2n.comparison}
\end{figure}

\begin{table}[tb]
\centering
\caption{Evaluation of our model on scene editing. Our model achieves similar performance to IN2N~\cite{in2n} and per-frame IP2P~\cite{ip2p} in terms of CLIP text-image similarity of the edited frames and output captions. As IP2P is independently editing views, it is inferior to the other methods in terms of consistency between the neighbouring frames and the edits applied to them. In terms of image quality (NIQE), our method outperforms the other 3D baseline, IN2N, by producing sharper and higher quality renderings. }
\resizebox{0.48\textwidth}{!}{
\begin{tabular}{lcccccc}
\hline
\textbf{Method} &
  \textbf{\begin{tabular}[c]{@{}c@{}}Txt-Img\\ Sim.$\uparrow$\end{tabular}} &
  \textbf{\begin{tabular}[c]{@{}c@{}}Frame\\ Sim.$\uparrow$\end{tabular}} &
  \textbf{\begin{tabular}[c]{@{}c@{}}Edit\\ Sim.$\uparrow$\end{tabular}} &
  \textbf{\begin{tabular}[c]{@{}c@{}}Image\\ Sim.$\uparrow$\end{tabular}} &
  \textbf{\begin{tabular}[c]{@{}c@{}}Edit\\ PSNR$\uparrow$\end{tabular}} &
  \textbf{NIQE$\downarrow$} \\ \hline
IP2P~\cite{ip2p} & 0.2770 & 0.9669 & 0.8082 & 0.9111 & 19.44 & 4.02 \\
IN2N~\cite{in2n} & 0.2683 & 0.9865 & 0.8822 & 0.8649 & 28.70 & 6.43 \\
Ours             & 0.2673 & 0.9876 & 0.8754 & 0.8910 & 31.01 & 5.53 \\ \hline
\end{tabular}
}
\label{tab:3d.editing.results}
\end{table}

\begin{figure}[t]
  \centering
   \includegraphics[width=1.0\linewidth]{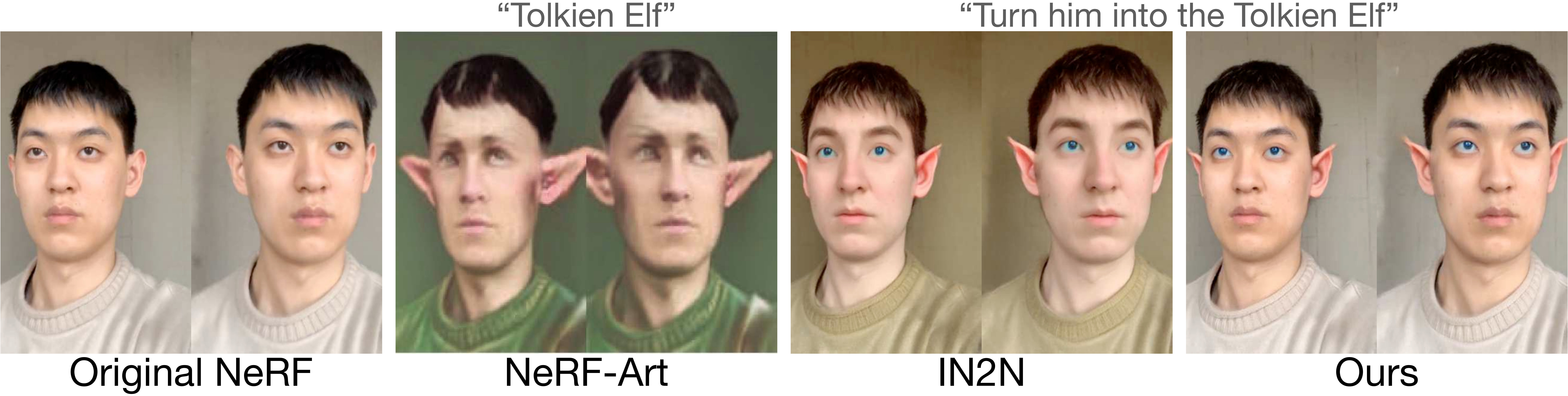}
   \caption{Comparison of our scene editing method against NeRF-Art~\cite{nerf.art} and IN2N~\cite{in2n}. 
   The baselines modify the background, shirt, and hair of the person, while our model only edits the eyes and ears.
   The extraneous changes of the baselines can even fail to preserve
   important scene semantics  (in this case, the individual's identity). 
   In contrast, our method applies only the minimum change required for the desired semantic alteration.
   }
   \label{fig:nerf.art.comparison}
\end{figure}

Qualitative examples of our scene editing results are shown in Figure~\ref{fig:3d.qualitative.results}. Our model edits the region most relevant to the edit, while keeping the rest of the scene unchanged. For example, while editing the bear statue and changing it to a \textit{panda}, a \textit{grizzly bear}, or a \textit{polar bear}, the background and the stage underneath the statue have remained intact, while the statue itself is changed to desired animals with sharp textures (notice the texture of the fur).

\begin{figure}[t]
  \centering
   \includegraphics[width=1.0\linewidth]{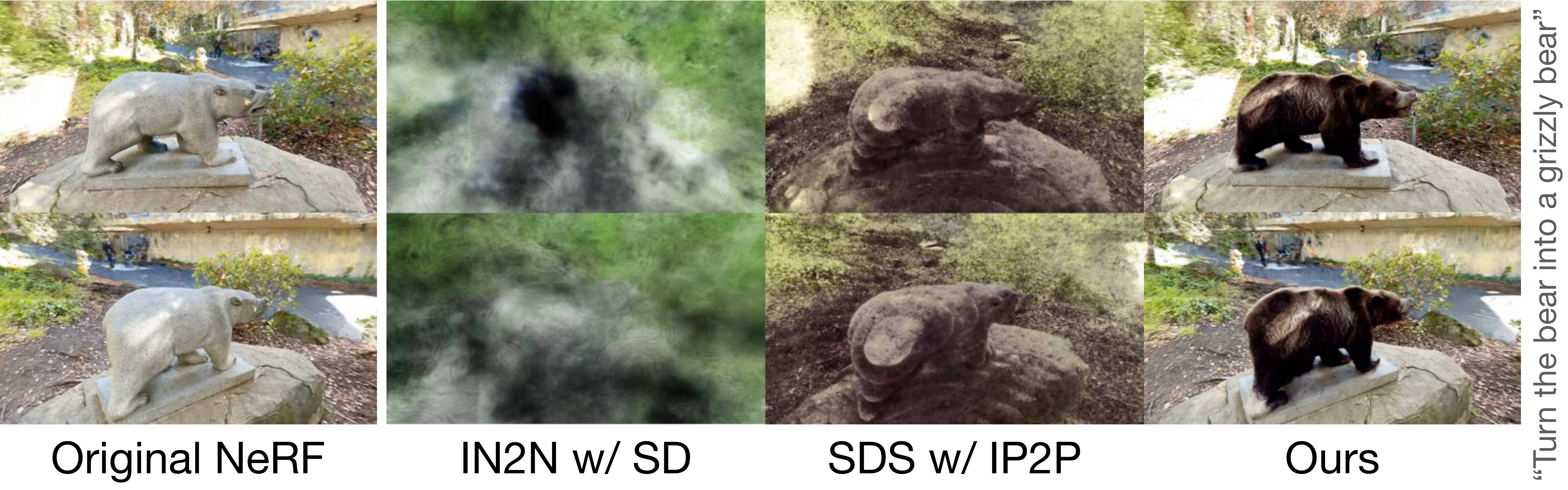}
   \caption{Qualitative comparison of our scene editing method against two baselines. \textit{IN2N w/ SD} performs the same iterative dataset updates as IN2N~\cite{in2n}, but with stable diffusion~\cite{stable.diffusion} instead of IP2P. \textit{SDS w/ IP2P} performs updates on the NeRF based on the SDS loss~\cite{poole2022dreamfusion} calculated via IP2P. Our method results in sharp outputs, while the baselines have failed on the task. }
   \label{fig:multiple.baselines.3d.comparison}
\end{figure}

We further provide qualitative comparisons to the baselines. As shown in Figure~\ref{fig:in2n.comparison}, built directly on IP2P, IN2N has the same tendency to over-edit scenes. In the case of giving \textit{the guy blue hair}, notice how it has also changed the t-shirt, eyes, and background colours. It has also changed the whole torso of the other person to a bronze statue, where the prompt has only asked for changing the face. In the bear scene, the background in IN2N output is blurred. This is due in part to the ambiguity of VAE decoder in upsampling, resulting in minor misalignments between different views. 
Moreover, since IP2P does not attempt to prevent changes to the background, some of the edited views have an altered background.
This inconsistency has resulted in a loss of sharpness. 
It also disrupts the optimization, as network capacities and loss gradients are allocated to background inconsistencies; hence, IN2N outputs are not as sharp as our result.
Moreover, IP2P is constrained to only edit the bear to a panda in our case, rather than trying to edit the whole image to satisfy the instruction. Consequently, the edited views in our method are more likely to be consistent, especially for nearby views, which is another reason that even our edited regions are considerably sharper, e.g., the texture of the panda's fur. 
In Figure~\ref{fig:nerf.art.comparison}, NeRF-Art~\cite{nerf.art} has followed the instruction and changed the face to the \textit{Tolkien Elf}, but the edited scene has quality artifacts associated with CLIP-based~\cite{clip.paper} methods, and has changed irrelevant regions of the scene, including the hair, background, and t-shirt. 
Figure~\ref{fig:multiple.baselines.3d.comparison} compares our method with additional baselines.

\begin{figure}[t]
  \centering
   \includegraphics[width=1.0\linewidth]{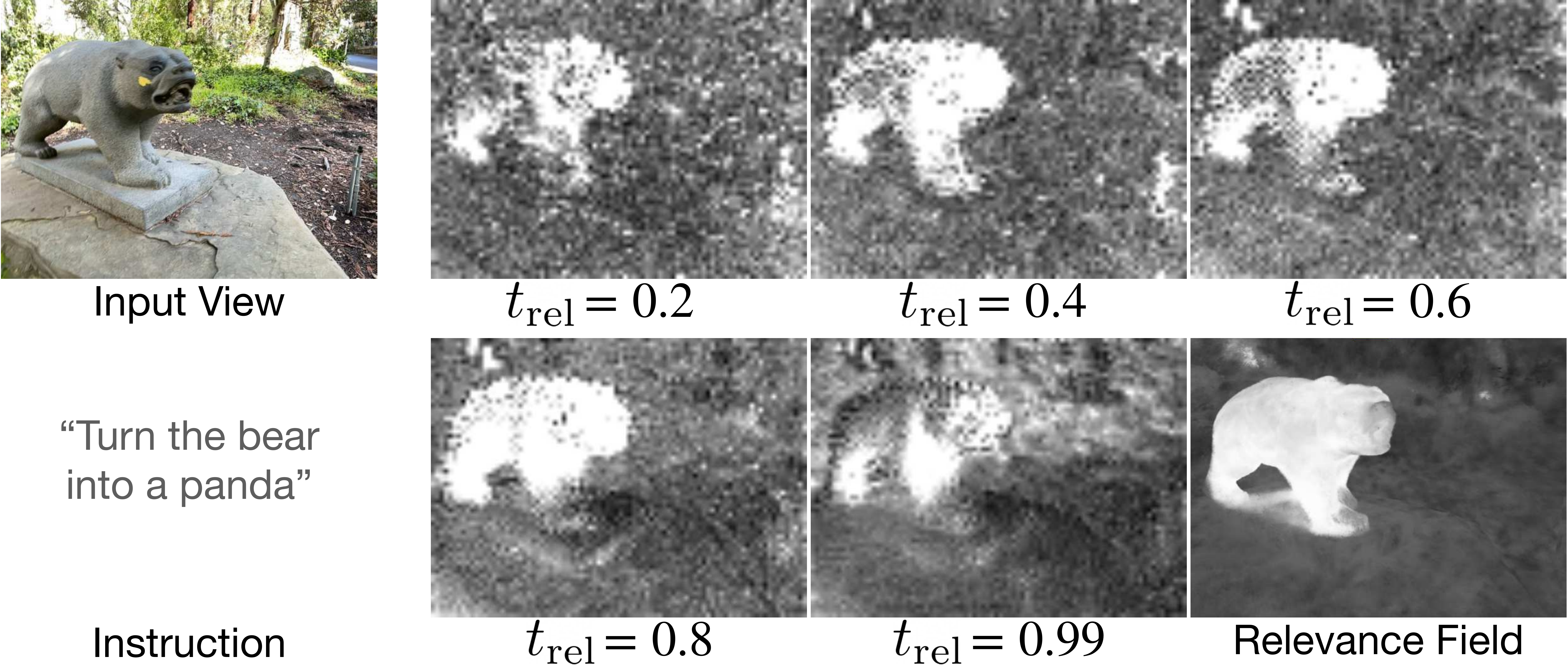}
   \caption{Comparison of the relevance maps calculated with different noise levels and a rendered relevance from a relevance field.  }
   \label{fig:relevance.map.ablation}
\end{figure}

\noindent\textbf{Relevance noise level.}
The relevance noise level, $t_\text{rel}$, is a hyperparameter we use for the calculation of the relevance field. Figure~\ref{fig:relevance.map.ablation} shows a comparison between the maps calculated using different noise levels. In our experiments, we found $t_\text{rel}=0.8$ to be reliable. 
This way, the relevance map is calculated using predictions in the higher-noise stages. As a result, the denoising process is fixating on the global structure of the generated images, rather than the fine details~\cite{balaji2022eDiff-I}. Thus, the predicted relevance masks encapsulate the global boundaries of the relevant regions. 
Moreover, Figure~\ref{fig:relevance.map.ablation} shows the rendering of the relevance field from the same view. 
Since the relevance field is supervised using maps from multiple views, it is effectively an ensemble over multiple predictions, and is more accurate than each single map. 
In addition to this ensemble nature,
the inductive bias of the NeRF architecture limits high-frequency field variations; hence, relevance renders provide a smooth consensus over the global scene structure, with minimal noise.

\begin{figure}[t]
  \centering
   \includegraphics[width=1.0\linewidth]{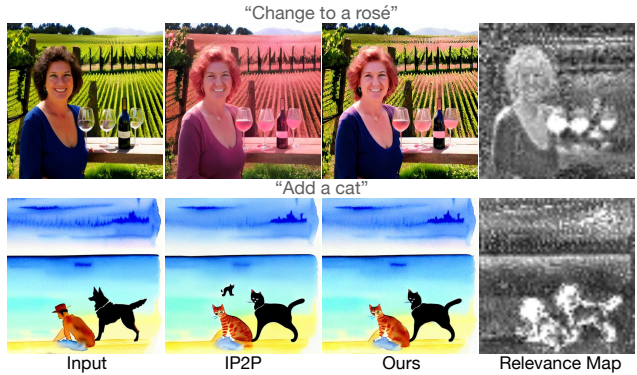}
   \caption{Examples of the failure cases of our image editing model. 
   Due to the reliance of our model on IP2P for predicting the relevance map and the edit, although our model is able to outperform IP2P by localizing the edits, it naturally cannot recover from IP2P's catastrophic failures. 
   }
   \label{fig:2d.failures}
\end{figure}

\noindent\textbf{Failure cases.}
The backbone of our method is a pre-trained IP2P~\cite{ip2p}. As a result, although our mask-guidance is able to alleviate the over-editing problem of IP2P, and to fix the upsampling ambiguity issue, it still is unable to recover from the cases in which IP2P fails badly. In Figure~\ref{fig:2d.failures}, we provide examples of such failures. For instance, in the first row, the prompt is \textit{``change to a rosé"}. Given the context of the image, the goal is to only change the drinks. However, IP2P has completely changed the field in the background and the hair colour to pink. This failure is reflected in the predicted relevance mask, which superfluously highlights those areas.
Although the result of our model is arguably better, it has still edited parts that were unnecessary to change. 
In the second example, \textit{``add a cat"}, localizing the edit with respect to the prompt is an ambiguous problem. The relevance map has failed to localize a certain position for the cat to be added, and instead, the person and the dog have been replaced with cats. 
Our method is agnostic to the underlying instruction-conditioned diffusion model, and can benefit from swapping IP2P with a better one in the future. 

\section{Conclusion}

We propose a method for predicting the relevance of each image pixel to an editing task based on a text instruction. This is done by looking at the discrepancy between a conditional and an unconditional pass over a diffusion-based image editor. We use this relevance as a mask to guide the generation and force the unmasked pixels to not change, resulting in a localized image editor.
We further show that training a relevance field on the relevance maps of the training views of a NeRF achieves similar localizations when editing 3D scenes. Our method shows superior performance compared to the baselines in both image and scene editing tasks.

\textbf{Acknowledgments}.
This work was conducted at Samsung AI Centre Toronto and it was funded by Mitacs and Samsung Research, Samsung Electronics Co., Ltd.

{\small
\bibliographystyle{ieee_fullname}
\bibliography{egbib}
}

\clearpage \newpage

\appendix

\section{Mask threshold effect}

\begin{figure}[tb]
  \centering
   \includegraphics[width=1.0\linewidth]{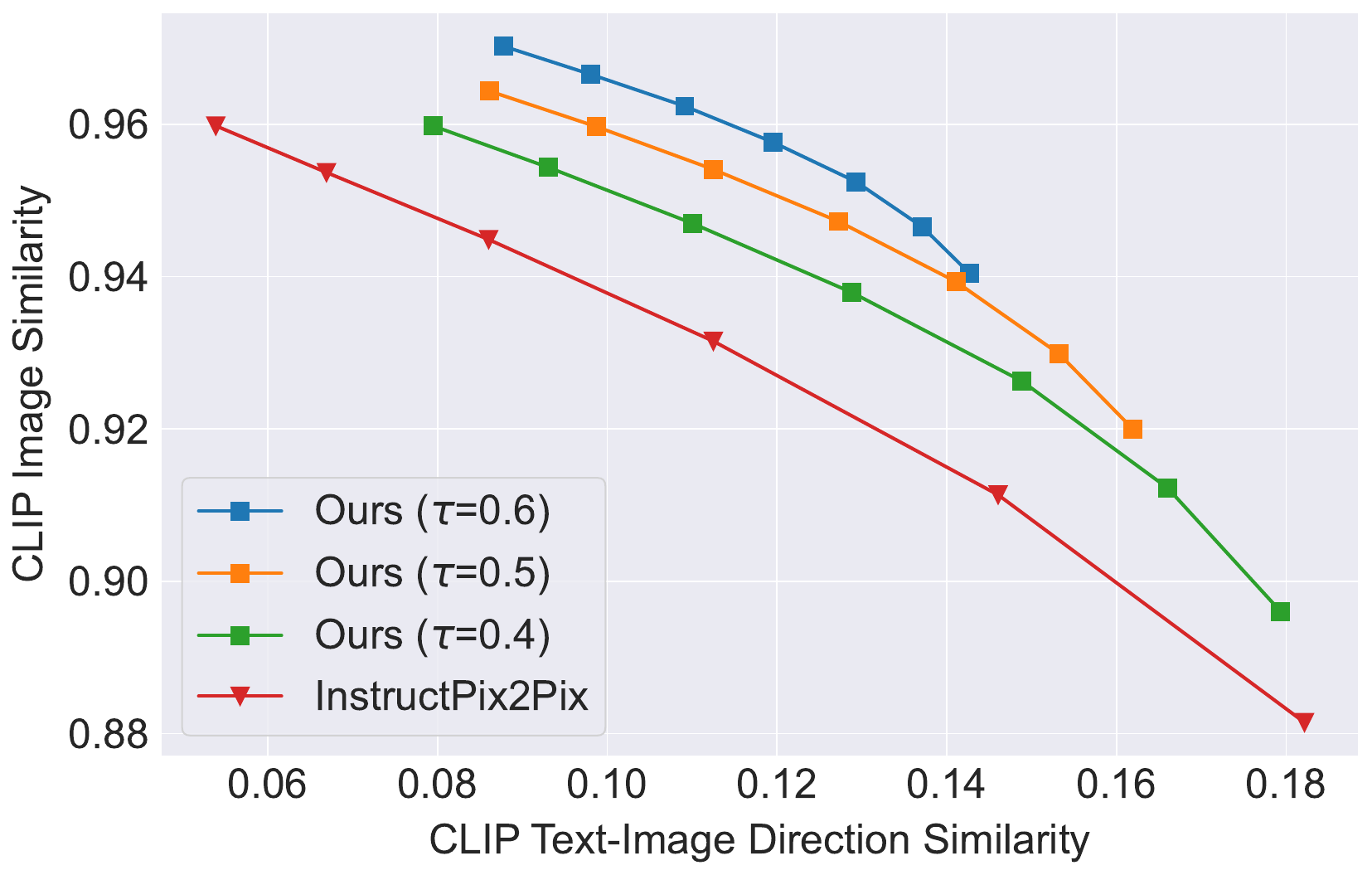}
   \caption{
   Quantitative comparison of our image editing method with different mask threshold values, $\tau$, against IP2P~\cite{ip2p}. 
   The text-guidance is set to $7.5$ for both method. For IP2P and our method, $S_I$ is changed between $[1.2, 2.2]$ and $[1.0, 2.2]$, respectively.  
   }
   \label{fig:2d.comparison.quantitative.fixed.tau}
\end{figure}

Figure~\ref{fig:2d.comparison.quantitative.fixed.tau} provides a quantitative comparison of our image editing method with different mask thresholds against IP2P~\cite{ip2p}. Our method produces outputs that closely reflect the desired edits (x-axis) while remaining consistent with the inputs (y-axis) by confining the edits in the relevant region. However, while increasing the mask threshold results in higher fidelity to the input, overly increasing it can prevent the model to edit the parts that actually matter; the lines in Figure~\ref{fig:2d.comparison.quantitative.fixed.tau} cover a smaller text-image direction similarity region as the mask threshold, $\tau$, is increased. Based on this experiment, we've typically set $\tau$ between $[0.4, 0.5]$ throughout the paper. 

\begin{figure}[t]
  \centering
   \includegraphics[width=0.99\linewidth]{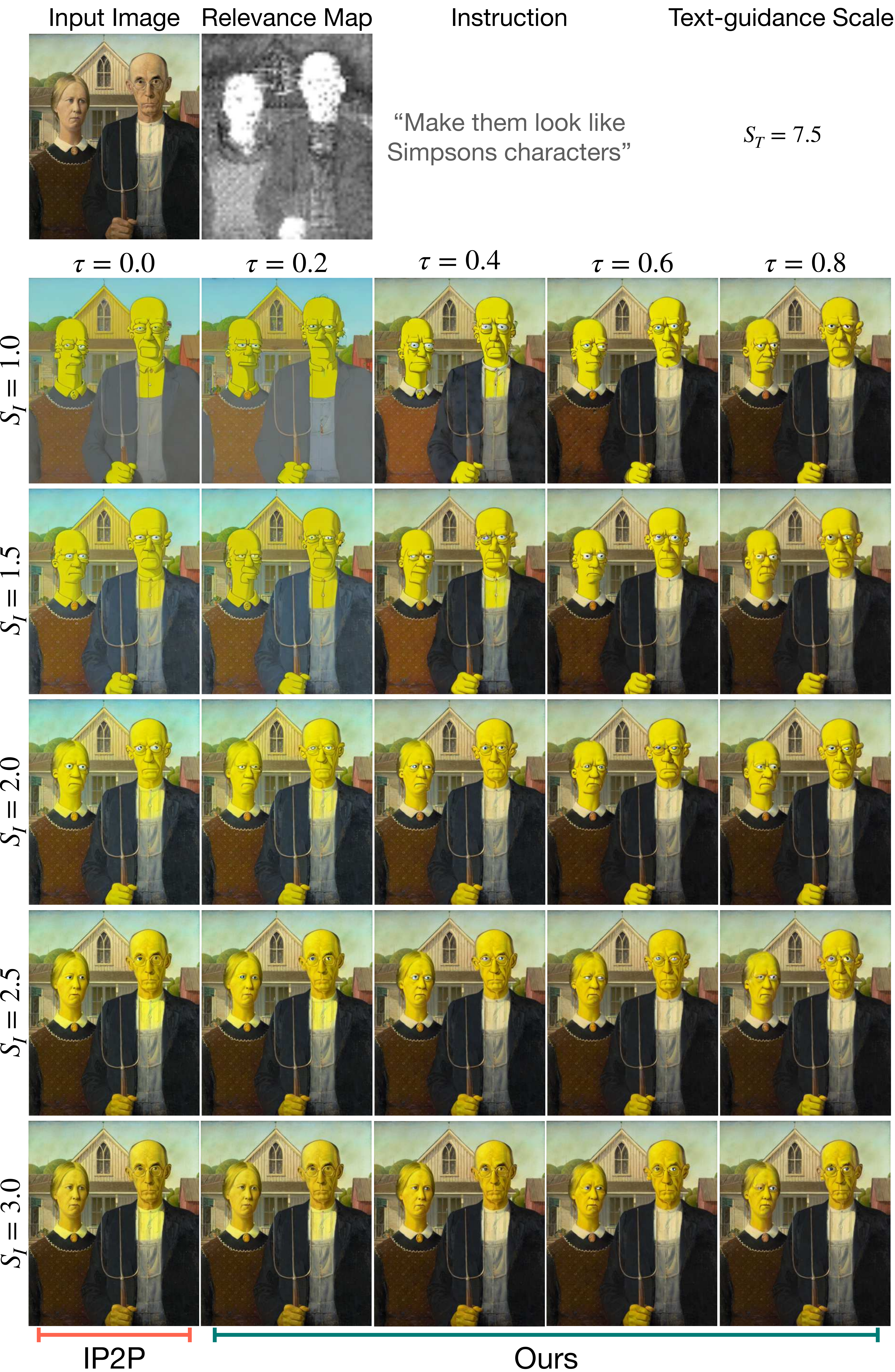}
   \caption{Qualitative comparison of the effects of the mask threshold, $\tau$, and the image guidance scale, $S_I$. Increasing the image guidance scale improves the similarity of the output and the input image, but significantly decreases the intensity of the edit, resulting in a failure. In contrast, $\tau$ provides a knob to the user to control the region of the edit, with reduced effects on the quality of the edit. }
   \label{fig:2d.different.knobs.comparison.simpsons}
\end{figure}

We further provide a qualitative example to showcase the effect of the mask threshold on the generation of the edited image (Figure~\ref{fig:2d.different.knobs.comparison.simpsons}). Setting the mask threshold, $\tau$, to $0$ results in every pixel to be masked. As a result, our model with $\tau=0$ is equivalent to IP2P~\cite{ip2p}.  For each $\tau$, we provide results with different image guidance scales, $S_I$. As evident in the results, in IP2P, simply increasing the image guidance scale is not enough to localize the edits; with $S_I=1.0$, the background and the clothes are drastically changed. When setting $S_I=3.0$ in IP2P, the woman's collar and the man's shirt are still changed to yellow, while the faces no more look like the Simpsons characters; increasing $S_I$ has an adverse effect on the text-image similarity, which is consistent with our quantitative findings in Figures~\ref{fig:2d.comparison.quantitative} and~\ref{fig:2d.comparison.quantitative.fixed.tau}. On the other hand, changing $\tau$ provides a different guidance knob to the user, and allows them to control the region to be edited, with a minimal damage to the regions that actually need to be modified. 

\section{Additional qualitative comparisons}

\begin{figure*}[t]
  \centering
   \includegraphics[width=0.99\linewidth]{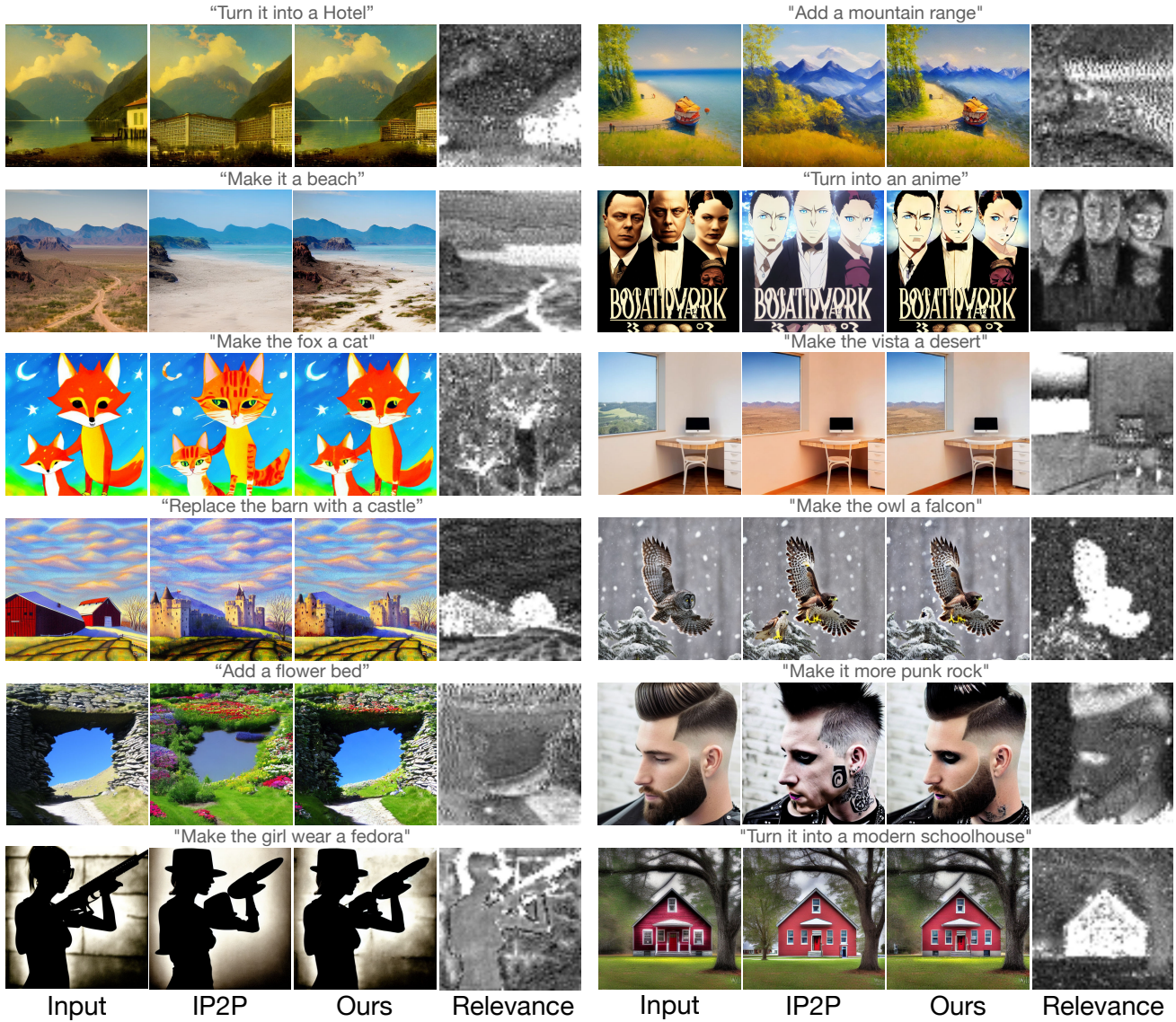}
   \caption{Additional Qualitative comparison of our image editing results against IP2P~\cite{ip2p}. For each image, we show the instruction, outputs of IP2P and our model, and the relevance map predicted by our model. Our model follows the instruction, while maintaining more consistency with the input image. This is due to the relevance-guidance, as only pixels with high relevance values are allowed to be modified. }
   \label{fig:ip2p.2d.comparison}
\end{figure*}

In Figure~\ref{fig:ip2p.2d.comparison}, we provide additional example to compare our localized image editing method with IP2P. In all of these examples, $S_I$ is between $[0.8, 1.0]$, and $S_T$ is always $7.5$. Mask thresholds are either $0.4$ or $0.5$. Foe each of the examples, we further provide the relevance map predicted by our method. Our results are more consistent with the input image by only locally changing the inputs in the regions with high relevance values. Meanwhile, our method has followed the instructions closely, and has resulted in images with similar or better edit qualities compared to IP2P.  

\section{Sample relevance field renderings}

\begin{figure*}[t]
  \centering
   \includegraphics[width=1.0\linewidth]{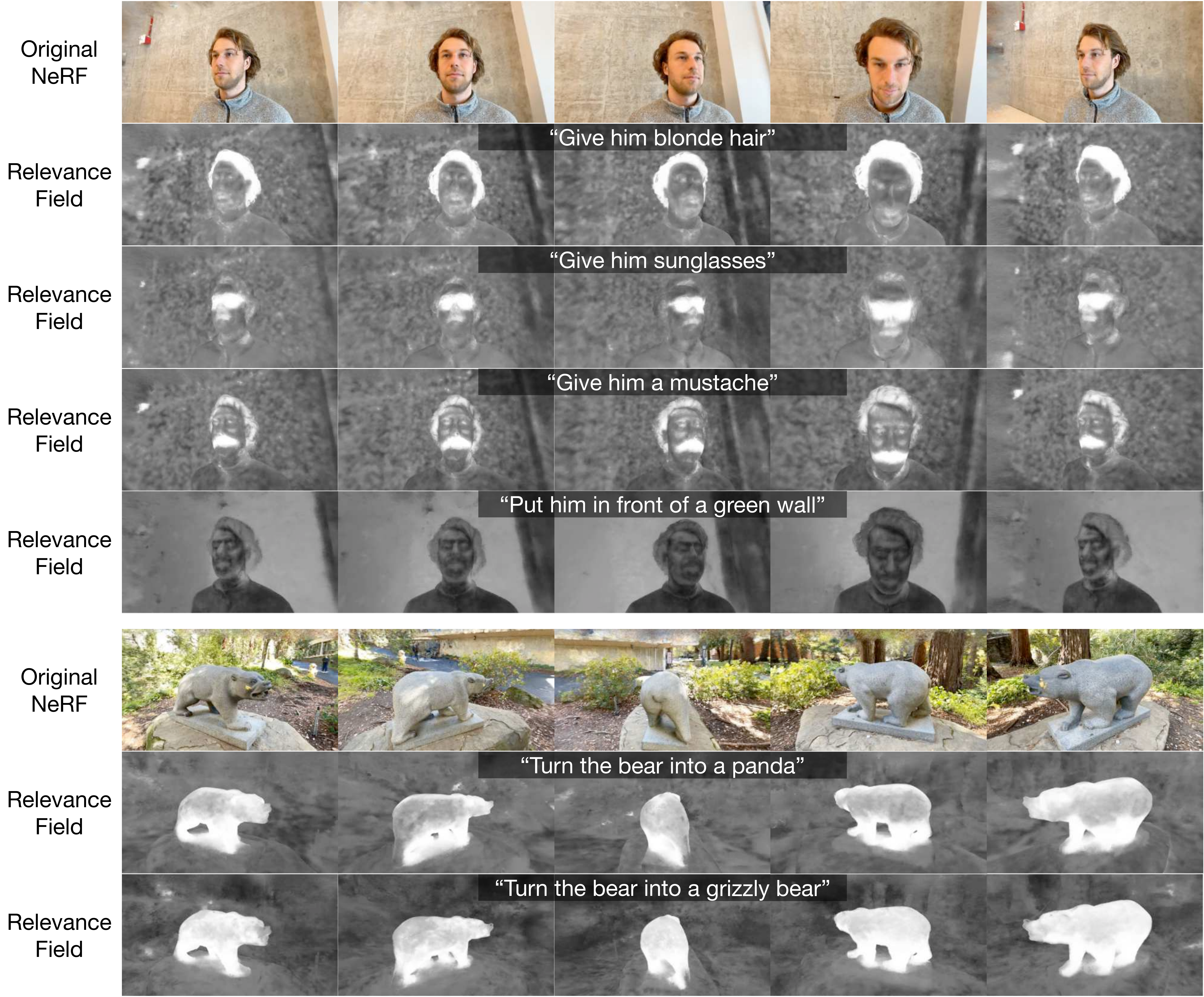}
   \caption{Sample rendered relevances from relevance fields trained on different scenes and different edit instructions. Each relevance field is visualized from multiple views, in addition to the corresponding views from the original NeRF model of the scene. Notice how each relevance field is mostly activated around the region that is highly relevant to the edit. For example, in the face scene and with the instruction \textit{``Give him blonde hair"}, only the hair is given high values in the field. This field allows to localize edits of the training views during each iterative update in a 3D consistent manner.  }
   \label{fig:sample.rendered.relevances}
\end{figure*}

In Figure~\ref{fig:sample.rendered.relevances}, we visualize sample relevance fields trained on different scenes and different edit instructions. Each relevance field is trained via 2D supervisions from relevance maps of the training views. As shown in the results, the relevance fields are mainly activated around the region that should be edited. To edit a NeRF, we use the rendered views from the relevance field as relevance-guidance for editing training views. As a result, the updates of training views during the iterative update process are only locally changed, and the changed region is consistent across different views. 

Note that the relevance field's densities (geometry) is always queried from the main NeRF model that is being edited. This is to ensure that potential inconsistencies between 2D relevance maps of different views does not hurt the main NeRF, and to enforce 2D relevance maps to be projected to the actual geometry of the scene and to become 3D-consistent; otherwise, the relevance field's geometry might converge to a degenerate solution to justify the inconsistencies. As the main NeRF is being updated towards the desired edited NeRF, its geometry might change. In that case, since the relevance field shares the same geometry, its 2D maps will be projected to the updated NeRF. Thus, during each update, the relevance field localizes the edits on the current version of the main NeRF. This allows slow changes in the geometry of the scene while only locally updating the views at each step, e.g., the addition of the mustache or the sunglasses, which require changes to the densities. 

\section{Additional Details}

For Figure~\ref{fig:more.2d.comparison.ip2p}, we set $S_T$ and $S_I$ to $7.5$ and $1$ respectively, while selecting $\tau$s proper to each edit. 
In Figure~\ref{fig:multiple.baselines.2d.comparison}, for our method, we set $S_T=7.5$, $S_I=0.8$, and $\tau=0.4$. For the other methods we used their default set of hyperparameters. In Figures~\ref{fig:2d.failures} and \ref{fig:ip2p.2d.comparison}, we set $S_T=7.5$, $S_I=0.8$ for both methods and $\tau=0.4$ for our method.
For NeRF editing experiments, $\tau$ is always set to $0.5$, and the guidance scales are as follows:
\begin{itemize}
    \item \textbf{Bear}: $S_T=6.5, S_I=1.5$
    \item \textbf{Face}: $S_T=7.5, S_I=1.5$
    \item \textbf{Farm}: $S_T=12.5, S_I=1.5$
    \item \textbf{Fangzhou}: $S_T=6.5, S_I=1.5$
\end{itemize}

\end{document}